\documentclass{article}

\usepackage{arxiv}

\usepackage[utf8]{inputenc} 
\usepackage[T1]{fontenc}    
\usepackage{hyperref}       
\usepackage{url}            
\usepackage{booktabs}       
\usepackage{amsfonts}       
\usepackage{nicefrac}       
\usepackage{microtype}      
\usepackage{xcolor}         
\usepackage{natbib}
\usepackage{times}
\usepackage{epsfig}
\usepackage{graphicx}
\usepackage{amsmath}
\usepackage{amssymb}
\usepackage{amsthm}
\usepackage{bbm}
\usepackage{microtype}
\usepackage{comment}
\usepackage{tabularray}
\usepackage{subcaption}

\usepackage{comment}

\usepackage{algorithm}
\usepackage{algpseudocode}
\usepackage{mathtools}
\usepackage{textcomp, gensymb}

\title{Good Actions Succeed, Bad Actions Generalize: A Case Study on Why RL Generalizes Better}

\author{
 Meng Song \\
 UC San Diego 
}
\date{}

\begin{document}
\maketitle

\begin{abstract}
Supervised learning (SL) and reinforcement learning (RL) are both widely used to train general-purpose agents for complex tasks, yet their generalization capabilities and underlying mechanisms are not yet fully understood. In this paper, we provide a direct comparison between SL and RL in terms of zero-shot generalization. Using the Habitat visual navigation task as a testbed, we evaluate Proximal Policy Optimization (PPO) and Behavior Cloning (BC) agents across two levels of generalization: state-goal pair generalization within seen environments and generalization to unseen environments. Our experiments show that PPO consistently outperforms BC across both zero-shot settings and performance metrics-success rate and SPL. Interestingly, even though additional optimal training data enables BC to match PPO's zero-shot performance in SPL, it still falls significantly behind in success rate. We attribute this to a fundamental difference in how models trained by these algorithms generalize: BC-trained models generalize by imitating successful trajectories, whereas TD-based RL-trained models generalize through combinatorial experience stitching-leveraging fragments of past trajectories (mostly failed ones) to construct solutions for new tasks. This allows RL to efficiently find solutions in vast state space and discover novel strategies beyond the scope of human knowledge. Besides providing empirical evidence and understanding, we also propose practical guidelines for improving the generalization capabilities of RL and SL through algorithm design.
\end{abstract}

\section{Introduction}
Supervised learning (SL) and reinforcement learning (RL) are two fundamental training paradigms for learning a general policy capable of solving various problems and diverse physical tasks. In the era of foundation models, both SL and RL play essential roles in guiding a general-purpose model to master specific downstream tasks. However, to what extent they can endow an agent's generalization ability and how they generalize remains unclear and has not been extensively explored \cite{gpt4, deepseek_r1, pi_0}.

To understand supervised learning and reinforcement learning's full capabilities in generalization, we train SL and RL agents from scratch and directly compare their zero-shot performance. This is different from prior work \cite{memo_gen} which runs RL on a model pretrained by SL. Although removing RL's dependence on SL leads to the failure of a general agent on most tasks, a few studies \cite{ddppo, autonomous_driving} have shown that with careful formulation of model inputs and reward functions, a pure RL agent can achieve strong zero-shot generalization in certain domains. In this paper, we use the Habitat visual navigation task as a testbed because it encompasses both geometric and visual variations while enabling the evaluation of both diverse and optimal solutions. We train Proximal Policy Optimization (PPO) \cite{ppo} and behavior cloning (BC) agents and evaluate their zero-shot generalization on two levels: {\it state-goal pair generalization}, which assesses the agent's ability to navigate between unseen start and goal pairs in training scenes, and {\it scene generalization}, which measures the agent's navigation ability in unseen scenes.

Our experiments show that PPO outperforms BC across both zero-shot evaluation settings and performance metrics—success rate in finding a path and success in finding the shortest path (measured by SPL). Notably, while augmenting BC with additional optimal training data can eliminate its performance gap with PPO in SPL, the gap in success rate persists. This suggests that PPO is trained to have the capability to always find a feasible solution, whereas BC specializes in finding a specific class of solutions. We suspect this arises from the distinct generalization mechanisms of TD-based RL algorithms and BC: RL generalizes by leveraging the failure trajectories while BC generalizes by imitating the success trajectories.

\begin{figure*}[t!]
        \centering
        \includegraphics[width=0.33\textwidth]{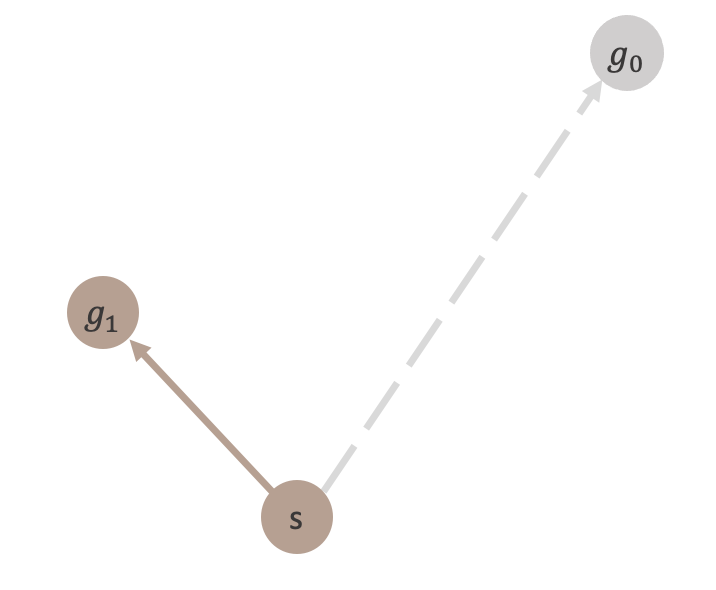}
    \caption{Trial-and-error data collection: The agent is commanded to reach $g_0$ but instead reaches $g_1$ either due to random exploration or the inability to reach $g_0$. Although these trajectories fail to accomplish the training tasks, they become useful for composing skills to solve unseen tasks.}
    \label{fig:trail_and_error_collection}
\end{figure*}

\begin{figure*}[t!]
        \centering
        \includegraphics[width=0.5\textwidth]{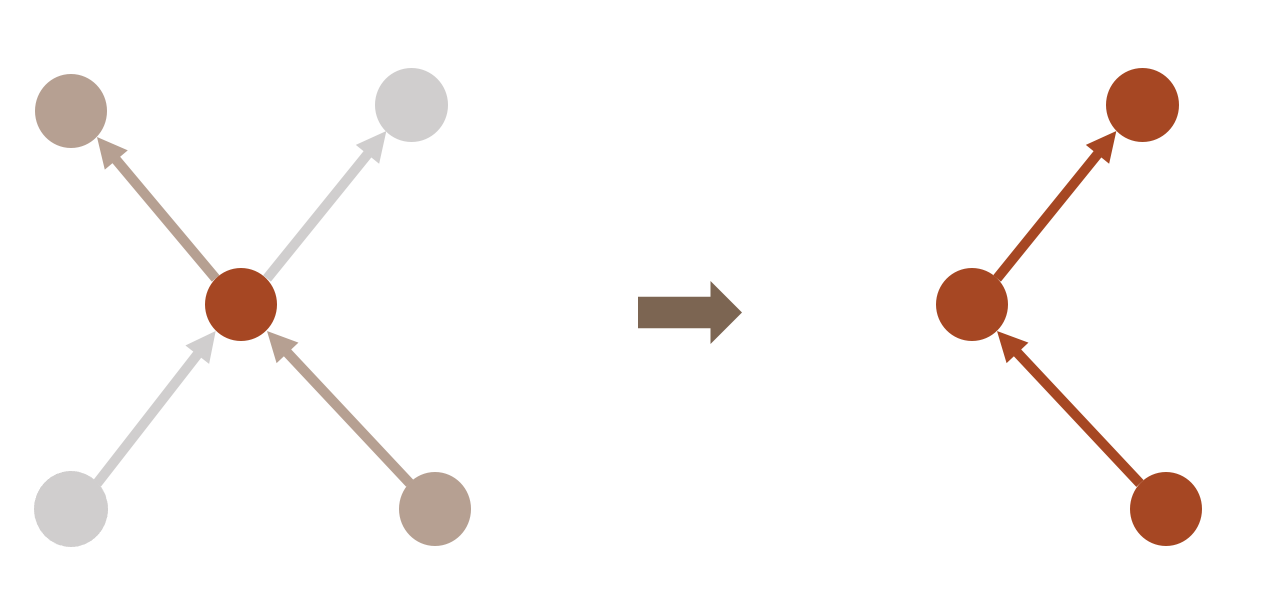}
    \caption{Combinatorial generalization: The agent has visited the gray and beige paths separately during training but has never seen the red path, yet it can discover it after TD learning.}
    \label{fig:comb_gen}
\end{figure*}

\begin{figure*}[t!]
        \centering
        \includegraphics[width=0.85\textwidth]{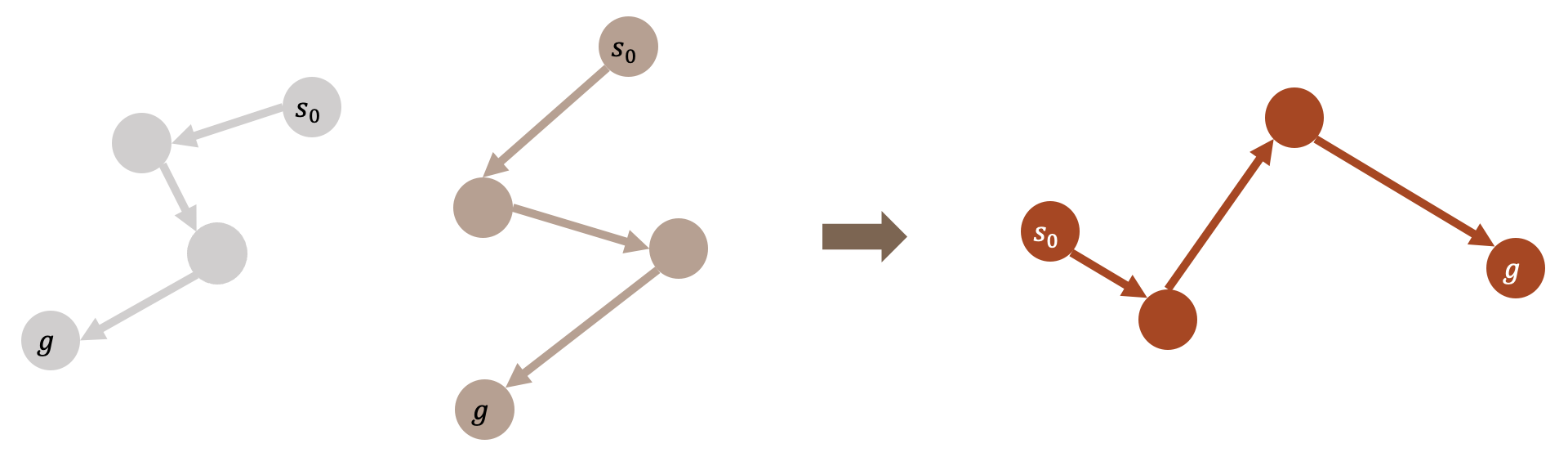}
    \caption{Feature generalization: The agent has been trained on a large set of optimal paths between different $(s_0,g)$ pairs. When presented with an unseen $(s_0,g)$, it infers the optimal path from $s_0$ to $g$ based on common features across training samples, such as the shape of optimal paths and frequently appearing decision-informative visual elements, etc.}
    \label{fig:feature_gen}
\end{figure*}

Specifically, RL solves unseen tasks by stitching together experiences \cite{ogbench, d4rl} collected during training, which are often suboptimal or failed trajectories for the original tasks (Figure \ref{fig:trail_and_error_collection}). Prior work \cite{td_sl} refers to this stitching phenomenon as {\it combinatorial generalization} (Figure \ref{fig:comb_gen}), which is attributed to the agent performing dynamic programming to optimize the training objective (i.e., TD learning). This notion of combinatorial generalization is completely different from how supervised learning (SL) generalizes. SL extracts features from the training data and generates an adapted solution for the unseen task. (Figure \ref{fig:feature_gen}). Its generalization is limited to the specific class of solutions defined by the training distribution. However, TD-based RL can compose an exponentially large set of possible solutions for the unseen task, which is particularly important for solving tasks in vast state space. This combinatorial nature of solution construction also holds the potential to creatively solve tasks that humans could never discover. In some sense, the distinction in the generalization behaviors of SL and RL is a reflection of the fundamental difference between the two definitions of AI: the ability to translate task instructions into human-like behaviors (SL), and the ability to acquire skills and discover solutions to achieve goals through interaction with the world (RL).

In summary, our primary contributions are:
\begin{itemize}
    \item We empirically demonstrate that a pure online RL algorithm can generalize better than BC, not only within a single training environment but also across unseen environments. 
    \item We find that although standard data augmentation can close the generalization gap between BC and PPO in terms of SPL, it fails to close the gap in success rate. We attribute this to their distinct generalization behaviors: TD-based RL generalizes by combinatorially stitching experiences, whereas BC generalizes by imitating the training samples. 
    \item Based on further analysis, we provide practical guidelines for improving the generalization capabilities of both RL and BC.  
\end{itemize}

\section{Habitat Visual Navigation Task}

\subsection{Task Definition}
We study the generalization abilities of SL and RL algorithms in the Habitat point goal visual navigation task \cite{habitat_challenge}. In this task, the agent is initialized at a random starting position and orientation in an indoor environment and asked to navigate to a target location specified relative to itself (e.g. ``Go 5m north, 10m east of you'') in the shortest path. At each timestep, in addition to this relative goal information, the agent receives an egocentric visual observation and takes an action from \{stop, move forward (0.25m), turn left ($10^\circ$), turn right ($10^\circ$)\}. An episode is considered successful if the agent issues a stop action within $0.2m$ of the goal position as measured by the geodesic distance. The episode terminates when either the episode length reaches 500 steps or the number of collision steps reaches 200.

\subsection{Goal Representation as Task Description}
Note that in the task definition, the goal is specified relative to the agent's current location rather than an absolute position. This design choice goes beyond a simple arithmetic operation or the question of whether the robot is equipped with GPS and a compass -- it carries a deeper significance. In fact, the subtle difference between absolute and relative goal representations leads to a distinctly different nature of the task. With an absolute goal, the task is framed as a typical goal-conditioned RL problem, which usually requires a large amount of early exploration and long-term planning. In contrast, with a relative goal, the task becomes more analogous to an instruction following problem, where the instruction is conveyed not through language but through distance and orientation measures. 

In a single-room navigation task, we observed that replacing the relative goal with an absolute goal is equivalent to removing the goal input, leading to a $40\%$ degradation in the success rate for a PPO agent. Relative goal representations are not exclusive to navigation tasks \cite{congnitive_mapping, learn_not_learn, nav_3d, virl}, many recent works across various robotics domains share a similar task specification in spirit. For example, the Decision Transformer \cite{dt} uses return-to-go as an input to measure task progress, while $\pi_0$ \cite{pi_0} trains a robot foundation model capable of solving a wide range of dexterous tasks by following language instructions. We therefore hypothesize that formulating the task as an instruction-following problem makes it easier to solve, scale, and generalize.

\section{Problem Formulation}
\subsection{RL Formulation}
In the Habitat visual navigation task, the egocentric visual observation makes the true environment state only partially observable. As a result, each scene is naturally formulated as a POMDP $M=(\mathcal{O}, \mathcal{A}, \mathcal{P}, \mathcal{R}, \rho, G, \gamma)$, where $\rho$ is the initial state distribution and $G$ is the goal state distribution. At each timestep $t$, the agent takes an action $a_t$ based on the observation history $h_{0:t}=(o_{0:t}, a_{0:t-1})$ and then receives the next observation $o_{t+1} \sim P(\cdot| h_{0:t}, a_t)$ and reward $r_{t+1} \sim R(\cdot| h_{0:t}, a_t)$.  The policy $\pi(a_t|h_{0:t}, g)$ and value function $V^\pi(h_{0:t}, g)$ can be instantiated using any memory-augmented model such as a recurrent model or a transformer.

Given a set of scenes (POMDPs) $\mathcal{M}=\{M_i\}_{i=1}^N$, we uniformly sample a specific scene $M_i$ from it at the beginning of each episode. We then draw an initial starting position and orientation $s_0 \sim \rho_i$ and a goal location $g \sim G_i$ to start the episode.

\subsection{Reward Function}
We follow the definition of the reward function in \cite{habitat_challenge, ddppo}, which is
\begin{equation}
    r_t = \left\{
\begin{array}{ll}
      s + d_{t-1} - d_{t} + \lambda \quad \text{if goal is reached}\\
      d_{t-1} - d_{t} + \lambda \quad \text{otherewise}\\
\end{array} 
\right. 
\end{equation}
where $s=2.5$ is the success reward, $\lambda = -0.0001$ is the time-elapse reward, and $d_{t-1} - d_{t}$ represents the change in geodesic distance to the goal from timestep $t-1$ to $t$. Computing the geodesic distance requires the ground-truth map during training but is unnecessary at the test time, as the agent does not rely on the reward at that stage.

In principle, the time-elapse reward $\lambda$ should encourage movement and shortest path discovery. However, the inaccurate value estimation limits its long-term effectiveness. Therefore, an additional distance-change reward is introduced to provide stepwise guidance on how each action contributes to the final goal. Our experiments show that removing this term causes the agent to fail completely.

\section{Algorithms}
We use Proximal Policy Optimization (PPO) \cite{ppo} and behavior cloning (BC) as example algorithms from RL and supervised learning to study their generalization and memorization abilities. We choose PPO over other model-free algorithms because \cite{ddppo} has shown that, by scaling up the training to 2.5 Billion steps, PPO can achieve near-perfect zero-shot performance on unseen environments in the evaluation task, making it a strong competitor.

We train both algorithms on $N$ training scenes and evaluate them on $M$ unseen scenes. In each training scene, we sample $K$ $(s_0,g)$ pairs for training and generate the optimal trajectory for each $(s_0,g)$ using the shortest path planner. 

Both PPO and BC policies are implemented as recurrent neural networks that condition on the entire history rather than a truncated context, as used in Transformers. 

\subsection{PPO}
\begin{figure*}[t!]
        \centering 
        \includegraphics[width=0.6\textwidth]{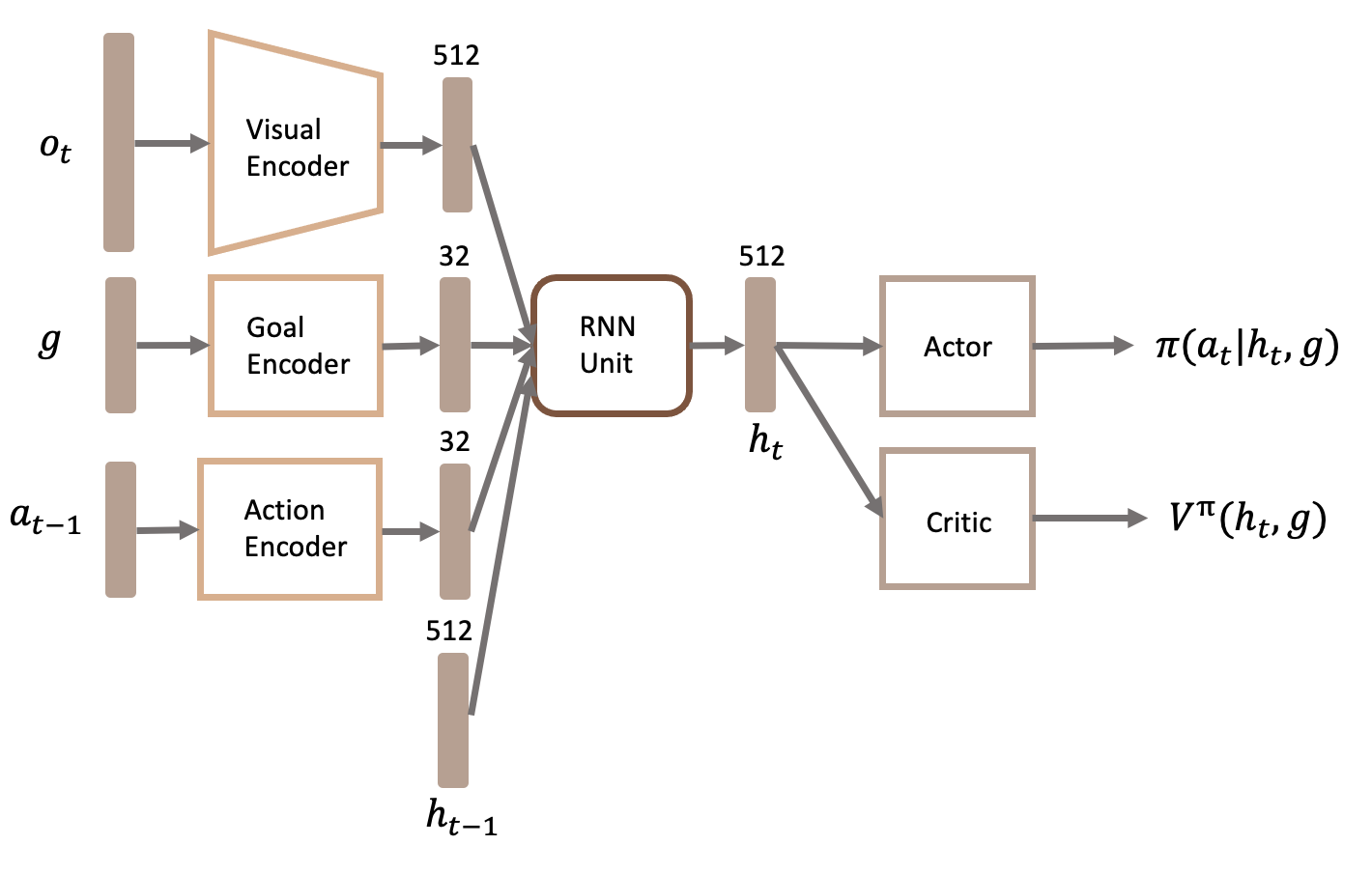}
        \caption{PPO Architecture}
    \label{fig:ppo_arch}
\end{figure*}

Given a training state-goal pair $(s_0,g)$ in scene $M$, PPO collects a set of trajectories 
$$\mathcal{D}_k=\{\tau_i=(o_0, a_0, o_1, r_1, \dots, a_{T_{i}}, o_{T_{i+1}}, r_{T_{i+1}})\}$$ 
by running $\pi_{\theta_k}$ in environment $M$, then updates the policy $\pi_\theta$ by maximizing the following objective:
\begin{equation}
 \theta_{k+1} = \arg \max_\theta \frac{1}{|\mathcal{D}_k|T_i}\sum_{\tau_i \in \mathcal{D}_k} \sum_{t=0}^{T_i} \mathcal{L}\left(h_{0:t}, a_t, \theta_{k}, \theta\right)
\end{equation}
\begin{equation}
    \mathcal{L}\left(h_{0:t}, a_t, \theta_{k}, \theta\right)=\min \Big(r_k(\theta) A^{\pi_{\theta_{k}}}(h_{0:t}, a_t, g), \quad \operatorname{clip}\left(r_k(\theta), 1-\epsilon, 1+\epsilon\right) A^{\pi_{\theta_{k}}}(h_{0:t}, a_t, g)\Big)
\end{equation}
where
\begin{equation}
    r_k(\theta) = \frac{\pi_{\theta}(a_t|h_{0:t}, g)}{\pi_{\theta_{k}}(a_t|h_{0:t}, g)}
\end{equation}
\begin{equation}
A^{\pi_{\theta_{k}}}(h_{0:t}, a_t, g)=R_t-V^{\pi_{\theta_{k}}}(h_{0:t},g)
\end{equation}
\begin{equation}
R_t = \sum_{j=t+1}^{T_{i+1}} \gamma^{j-t} r_j
\end{equation}
We follow the architecture design in \cite{ddppo}, which implements the recurrent policy (actor) $\pi(a_t|h_{0:t}, g)$ and the recurrent value function (critic) $V^\pi(h_{0:t}, g)$ using a shared RNN encoder (Figure \ref{fig:ppo_arch}). 

\subsection{Behavior Cloning}
\begin{figure*}[t!]
        \centering 
        \includegraphics[width=0.6\textwidth]{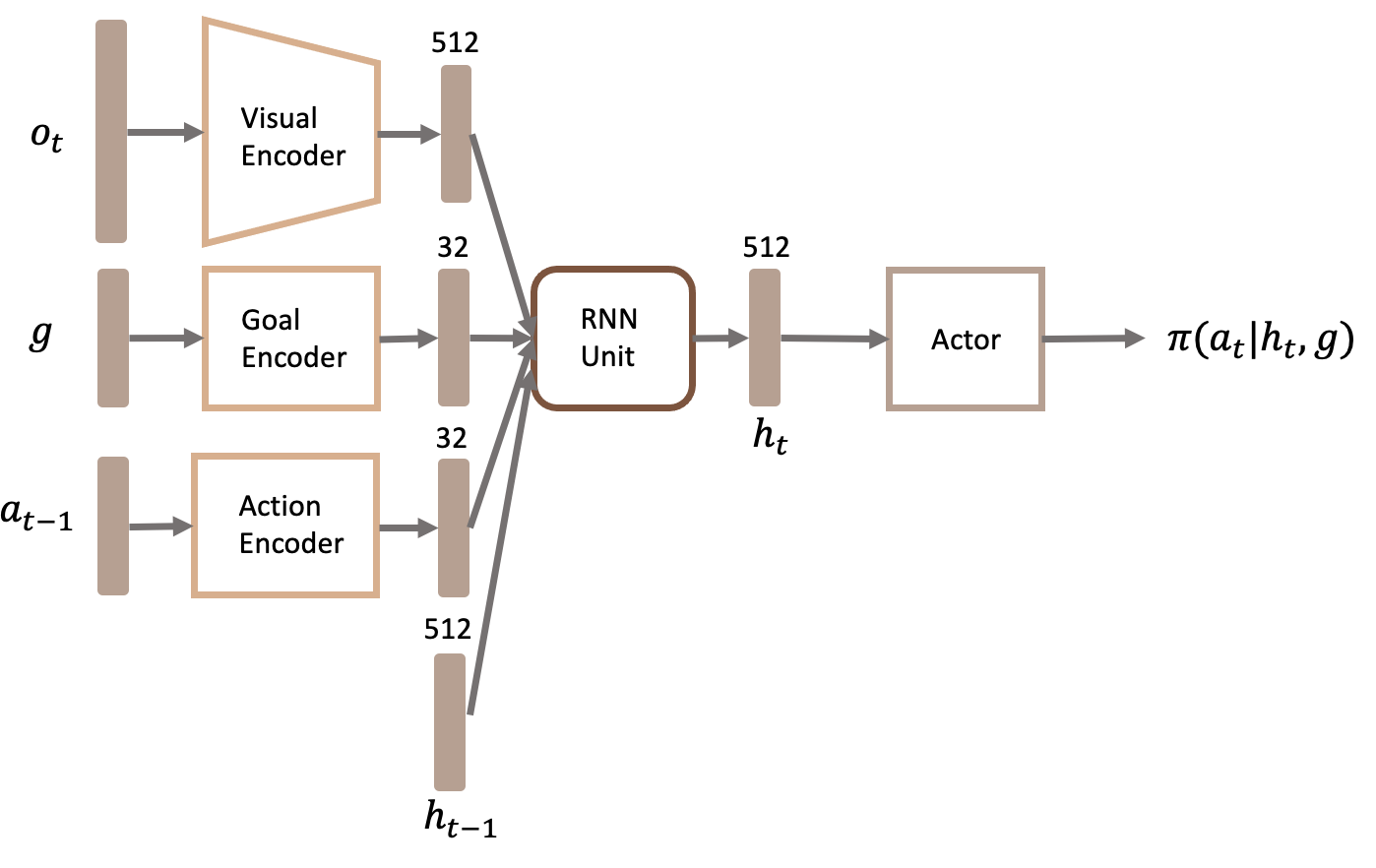}
        \caption{BC Architecture}
    \label{fig:bc_arch}
\end{figure*}

Given a set of optimal training trajectories 
$$\mathcal{D}_{\text{train}}=\{\tau_i=(o^*_0, a^*_0, o^*_1, \dots, a^*_{T_{i}}, o^*_{T_{i+1}})\}$$ where each $\tau_i$ has an associated goal $g_i$, 
the learning objective of behavior cloning (BC) can be written as
\begin{equation}
    \arg \max_\theta \frac{1}{|\mathcal{D}_{\text{train}}|T_i} \sum_{\tau_i \in \mathcal{D}_{\text{train}}} \sum_{t=0}^{T_i} \ln \pi_{\theta}(a^*_t|h^*_{0:t}, g_i)
\end{equation}
where 
$$
h^*_{0:t}=(o^*_{0:t}, a^*_{0:t-1})
$$
To ensure a fair comparison, the BC agent employs the same architecture as the actor in PPO, but omits the critic (Figure \ref{fig:bc_arch}).

\section{Experiments}
\subsection{Experimental Setup}
We conduct experiments using the training and validation sets of the Gibson dataset \cite{gibson}. The agents are trained on a subset of 4 scenes from the training set and evaluated on all 14 scenes in the validation set. The evaluation comprises three settings:
\begin{itemize}
    \item {\bf Seen $(s_0,g)$}:
    Evaluate the agent's memorization ability on the seen $(s_0,g)$ pairs from the training scenes. 
    \item {\bf Unseen $(s_0,g)$}:
    Evaluate the agent's zero-shot generalization ability on unseen $(s_0,g)$ pairs from the training scenes. This setting specifically assesses the agent's geometric generalization ability, which refers to its capability to infer paths between new locations in previously seen environments. This process primarily relies on the agent's understanding and reasoning of geometric information such as location, distance, layout, and spatial structures.
    
    \item {\bf Unseen scenes}: 
    Evaluate the agent's zero-shot generalization ability in unseen scenes. In this setting, the agent needs to also demonstrate visual generalization—the ability to transfer learned policies to a new environment based on visual understanding. For example, the agent should learn to move forward when a hallway is ahead and turn when a wall is in front, regardless of variations in appearance. It must then apply these learned skills in an unseen environment with new spatial layouts and visual features. 
\end{itemize}

During evaluation, to fully compare the learned policies, we sample actions from the policy distribution of PPO and BC instead of greedily selecting the most probable ones.

\subsection{Datasets}
Based on the experimental setup, we construct the training and evaluation datasets as follows: We sample 2000 $(s_0,g)$ pairs from each training scene for training and a separate set of 200 pairs for unseen $(s_0,g)$ evaluation. Within the 2000 training pairs, we sample 200 pairs for seen $(s_0,g)$ evaluation. For the unseen scenes evaluation, we use the full validation set, which consists of 994 $(s_0,g)$ pairs from 14 scenes.

During training, for each $(s_0,g)$ pair, the PPO agent learns from trajectories collected under its current policy, while BC is trained on the optimal trajectories generated by the shortest path planner.

\subsection{Evaluation Metrics}
The agent's performance is measured using the official evaluation metrics: success rate and SPL, where SPL refers to the success weighted by path length metric, defined as follows:
\begin{equation}
    \mathrm{SPL}=\frac{1}{N} \sum_{i=1}^{N} S_{i} \frac{l_{i}}{\max \left(p_{i}, l_{i}\right)}
\end{equation}
where for the i-th episode, $l_i$ is the length of the shortest path between the start
and goal, $p_i$ is the length of the agent’s path, and $S_i$ is the binary indicator of success. $SPL=1$ only when the agent's path exactly matches the ground truth shortest path.

Note that the success rate measures the agent's ability to find any path to the goal, allowing multiple solutions to the task, while SPL evaluates the ability to find the unique shortest path to the goal, which is more strict. The distinction of these two metrics helps us identify the different generalization strategies of BC and PPO in solving zero-shot tasks. PPO leverages its combinatorial generalization ability to efficiently compose diverse possible paths, increasing the likelihood of reaching new goals and thereby achieving a high success rate. In contrast, BC excels at extracting common patterns from the training shortest paths and generalizing based on these patterns, allowing it to attain a high SPL.

\section{Results}
We conduct experiments to compare and analyze the generalization performance of BC and PPO agents by investigating the following questions:

\subsection{How do BC and PPO agents perform in both training and zero-shot tasks?}
\begin{figure*}[t!]
        \centering
        \begin{subfigure}{0.6\textwidth} 
        \centering
        \includegraphics[width=\textwidth]{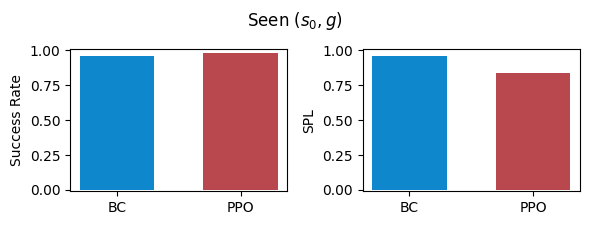}
        \end{subfigure}
        \centering
        \begin{subfigure}{0.6\textwidth} 
        \centering
        \includegraphics[width=\textwidth]{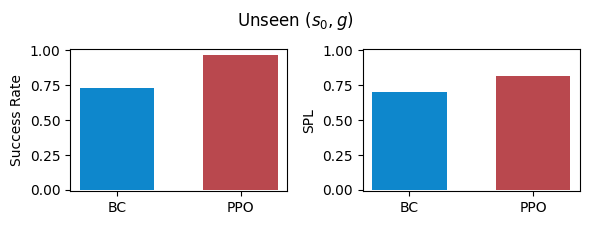}
        \end{subfigure}
        \centering
        \begin{subfigure}{0.6\textwidth} 
        \centering
        \includegraphics[width=\textwidth]{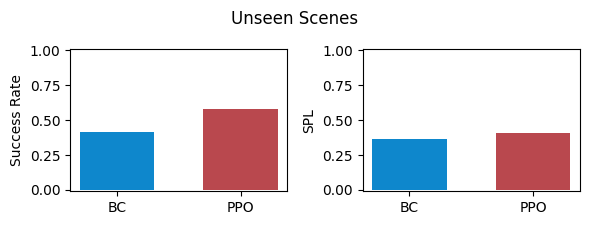}
        \end{subfigure}
    \caption{Performance comparison of BC and PPO on the training and zero-shot tasks.}
    \label{fig:compr}
\end{figure*}

As illustrated in Figure \ref{fig:compr}, when evaluating on seen $(s_0, g)$ pairs, we observe that BC achieves about $96\%$ SPL and success rate, outperforming PPO by $13\%$ in SPL, but degrades in unseen tasks. In contrast, PPO consistently outperforms BC in both zero-shot tasks and across two evaluation metrics, especially when generalizing to unseen $(s_0, g)$ pairs and when evaluated using success rate instead of SPL. In this case, PPO achieves a $96.75\%$ zero-shot success rate.

This suggests that BC is good at memorization, while PPO excels at generalization. In particular, PPO generalizes better in finding any feasible path to the goal than the shortest path, as its performance advantage is greater when measured by success rate than by SPL. It is worth noting that these observations hold not only to training scenes (MDPs), but also unseen scenes (MDPs), which demonstrates the broadness of RL's generalization capabilities.

\subsection{Is data augmentation able to close the gap between BC and PPO in zero-shot generalization?}
As the generalization abilities of PPO and BC rely on different training data — PPO trains on trajectories collected by its behavior policy, while BC trains on a static set of expert demonstrations, which are of high quality but of lesser amount — a natural question to ask is: Can we improve BC's generalization through data augmentation?

\begin{figure*}[t!]
        \centering
        \begin{subfigure}{0.6\textwidth} 
        \centering
        \includegraphics[width=\textwidth]{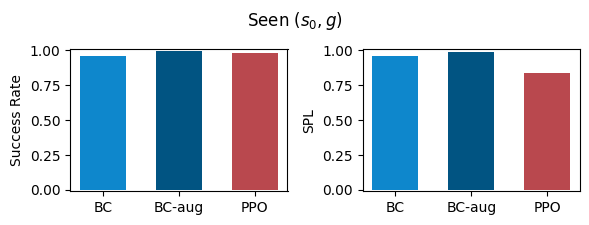}
        \end{subfigure}
        \centering
        \begin{subfigure}{0.6\textwidth} 
        \centering
        \includegraphics[width=\textwidth]{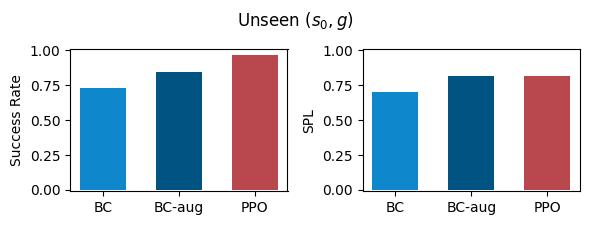}
        \end{subfigure}
        \centering
        \begin{subfigure}{0.6\textwidth} 
        \centering
        \includegraphics[width=\textwidth]{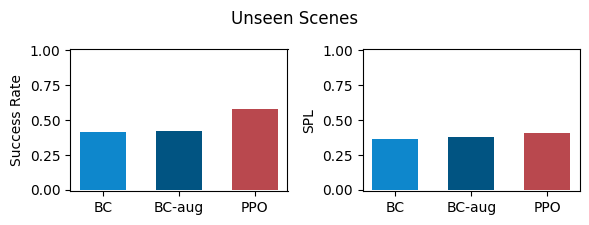}
        \end{subfigure}
    \caption{Performance comparison of BC, augmented BC, and PPO on the training and zero-shot tasks. The augmented BC is trained on 2000 additional optimal trajectories per scene.}
    \label{fig:aug_bc}
\end{figure*}

To answer this question, we augment BC's training data with the optimal trajectories of 2000 additional $(s_0, g)$ pairs in each training scene. These pairs do not overlap with either the original training pairs or the unseen $(s_0, g)$ pairs. As shown in Figure \ref{fig:aug_bc}, we observe that with more training demonstrations, the augmented BC achieves even better memorization performance, reaching nearly $100\%$ on seen $(s_0, g)$ pairs. Additional training data also improves BC's zero-shot performance.

Interestingly, in the unseen $(s_0,g)$ task and unseen scene task, the augmented BC still falls significantly behind PPO in success rate but nearly matches its performance in terms of SPL. In other words, there are always $(s_0,g)$ pairs that PPO can successfully navigate, but augmented BC cannot. Moreover, note that PPO's SPL is lower than its success rate, indicating that the path it finds is not always the shortest. In contrast, BC and augmented BC have equal SPL and success rate, meaning the paths they find are always the shortest.

We hypothesize that this phenomenon can be attributed to the distinct ways in which PPO and BC generalize. BC generalizes by learning the common patterns in training data—specifically, the shortest paths in our case. As a result, training with sufficient shortest path demonstrations allows it to achieve performance comparable to PPO in terms of SPL. In contrast, PPO generalizes by recombining pieces of trajectories into any possible paths between new $(s_0,g)$. Since most of these paths do not share common features with the shortest paths and are combinatorially numerous, PPO's advantage in success rate will always remain, even if BC is trained on an infinite number of optimal demonstrations.

\subsection{Does training in more scenes necessarily improve the performance?}
\begin{figure*}[t!]
        \centering
        \begin{subfigure}{0.6\textwidth} 
        \centering
        \includegraphics[width=\textwidth]{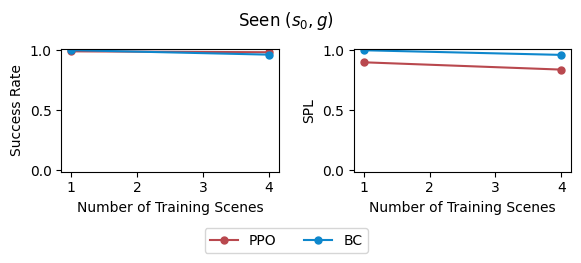}
        \end{subfigure}
        \centering
        \begin{subfigure}{0.6\textwidth} 
        \centering
        \includegraphics[width=\textwidth]{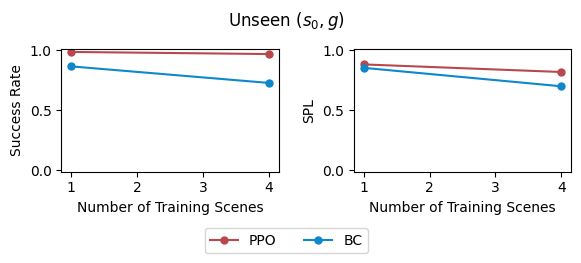}
        \end{subfigure}
        \centering
        \begin{subfigure}{0.6\textwidth} 
        \centering
        \includegraphics[width=\textwidth]{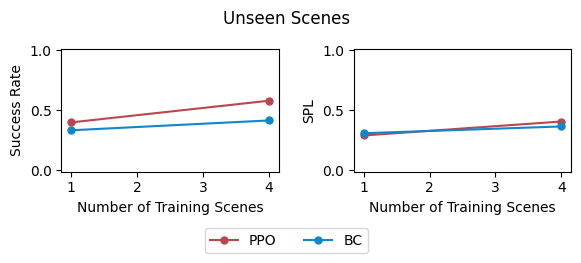}
        \end{subfigure}
    \caption{Performance change of BC and PPO as the number of training scenes increases from 1 to 4, evaluated on both the training and zero-shot tasks.}
    \label{fig:ablate_scenes}
\end{figure*}

To understand the impact of the number of training scenes on generalization, we train PPO and BC on 1 and 4 scenes respectively, and plot their evaluation performance in Figure \ref{fig:ablate_scenes}. Despite the variation in the number of training scenes, the experiments consistently show that BC outperforms PPO in memorization, while PPO outperforms BC in generalization, especially in finding any feasible paths to the goal.  

Notably, for both PPO and BC, training on more scenes facilitates the generalization to unseen scenes, but reduces the generalization performance on unseen $(s_0, g)$ and memorization performance on seen $(s_0, g)$. The performance decline is particularly noticeable for BC. We suspect that this is because increasing the number of training scenes improves the agent's visual understanding and generalization ability while simultaneously compromising its geometric generalization and memorization abilities. One can imagine a scenario where similar $(s_0, g)$ pairs exist in different scenes, but due to variations in layouts, their optimal paths differ. This conflicting information introduces confusion in the agent's memory, making it difficult for the agent to generalize correctly to new situations. Scaling up the model and increasing the training data—both in $(s_0,g)$ pairs and scenes—could potentially help reconcile the trade-off between visual and geometric generalization. 

\section{Conclusion and Discussion}
In this paper, we investigate the generalization behaviors of TD-learning-based reinforcement learning (RL) and supervised learning (SL) and how their distinct behaviors lead to generalization gaps. We explore this problem in the Habitat visual navigation task by training PPO and BC agents as instantiations. We evaluate their generalization abilities in two zero-shot settings: same-scene $(s_0,g)$ pair generalization and across-scene generalization, and using two evaluation metrics: SPL and success rate. Our results show that PPO exhibits superior generalization compared to BC across both zero-shot settings and evaluation metrics. 

To understand the reasons behind this phenomenon, we conduct further analysis. Our experiments suggest that PPO significantly outperforms BC in zero-shot success rate because it can reuse and recombine previously seen paths into a combinatorial number of solutions for new tasks. This unique generalization behavior stems from the dynamic programming nature of TD learning and the acquisition of non-optimal experiences through trial-and-error—capabilities that BC inherently lacks. However, BC follows a distinct generalization strategy, effectively extracting common patterns from the training trajectories and generating solutions that follow similar patterns to solve unseen tasks. This ability enables it to match PPO's zero-shot performance in finding the specific set of solutions (such as the shortest path in our case) when trained on a larger dataset, but not other possible solutions. 

By identifying the different generalization behaviors of RL and SL, we further provide practical suggestions on how generalization can be improved. On one hand, based on the combinatorial generalization of TD-learning-based RL methods, we suggest incorporating a maximum entropy regularization term into the regular RL objective to further improve its generalization ability. Maximum entropy RL preserves all possible solutions to the training tasks rather than over-committing to a single solution, potentially offering greater capacity for combinatorial generalization. On the other hand, supervised learning methods need to be trained on data of varying quality, including noisy and sub-optimal trajectories, to generate diverse solutions for new tasks and improve the success rate. To this end, both RL and SL policy should be multimodal to allow sufficient expressiveness. We also encourage future work on designing new algorithms that combine the best of both worlds—composing diverse solutions for new tasks while adapting learned patterns to novel cases.

\bibliographystyle{plainnat}
\bibliography{references}

\begin{thebibliography}{17}
\providecommand{\natexlab}[1]{#1}
\providecommand{\url}[1]{\texttt{#1}}
\expandafter\ifx\csname urlstyle\endcsname\relax
  \providecommand{\doi}[1]{doi: #1}\else
  \providecommand{\doi}{doi: \begingroup \urlstyle{rm}\Url}\fi

\bibitem[Black et~al.(2024)Black, Brown, Driess, Esmail, Equi, Finn, Fusai, Groom, Hausman, Ichter, Jakubczak, Jones, Ke, Levine, Li-Bell, Mothukuri, Nair, Pertsch, Shi, Tanner, Vuong, Walling, Wang, and Zhilinsky]{pi_0}
Kevin Black, Noah Brown, Danny Driess, Adnan Esmail, Michael Equi, Chelsea Finn, Niccolo Fusai, Lachy Groom, Karol Hausman, Brian Ichter, Szymon Jakubczak, Tim Jones, Liyiming Ke, Sergey Levine, Adrian Li-Bell, Mohith Mothukuri, Suraj Nair, Karl Pertsch, Lucy~Xiaoyang Shi, James Tanner, Quan Vuong, Anna Walling, Haohuan Wang, and Ury Zhilinsky.
\newblock $\pi_0$: A vision-language-action flow model for general robot control, 2024.
\newblock URL \url{https://arxiv.org/abs/2410.24164}.

\bibitem[Chen et~al.(2021)Chen, Lu, Rajeswaran, Lee, Grover, Laskin, Abbeel, Srinivas, and Mordatch]{dt}
Lili Chen, Kevin Lu, Aravind Rajeswaran, Kimin Lee, Aditya Grover, Michael Laskin, Pieter Abbeel, Aravind Srinivas, and Igor Mordatch.
\newblock Decision transformer: Reinforcement learning via sequence modeling, 2021.
\newblock URL \url{https://arxiv.org/abs/2106.01345}.

\bibitem[Chu et~al.(2025)Chu, Zhai, Yang, Tong, Xie, Schuurmans, Le, Levine, and Ma]{memo_gen}
Tianzhe Chu, Yuexiang Zhai, Jihan Yang, Shengbang Tong, Saining Xie, Dale Schuurmans, Quoc~V. Le, Sergey Levine, and Yi~Ma.
\newblock Sft memorizes, rl generalizes: A comparative study of foundation model post-training, 2025.
\newblock URL \url{https://arxiv.org/abs/2501.17161}.

\bibitem[Cusumano-Towner et~al.(2025)Cusumano-Towner, Hafner, Hertzberg, Huval, Petrenko, Vinitsky, Wijmans, Killian, Bowers, Sener, Krähenbühl, and Koltun]{autonomous_driving}
Marco Cusumano-Towner, David Hafner, Alex Hertzberg, Brody Huval, Aleksei Petrenko, Eugene Vinitsky, Erik Wijmans, Taylor Killian, Stuart Bowers, Ozan Sener, Philipp Krähenbühl, and Vladlen Koltun.
\newblock Robust autonomy emerges from self-play, 2025.
\newblock URL \url{https://arxiv.org/abs/2502.03349}.

\bibitem[DeepSeek-AI(2025)]{deepseek_r1}
DeepSeek-AI.
\newblock Deepseek-r1: Incentivizing reasoning capability in llms via reinforcement learning, 2025.
\newblock URL \url{https://arxiv.org/abs/2501.12948}.

\bibitem[Fu et~al.(2021)Fu, Kumar, Nachum, Tucker, and Levine]{d4rl}
Justin Fu, Aviral Kumar, Ofir Nachum, George Tucker, and Sergey Levine.
\newblock D4rl: Datasets for deep data-driven reinforcement learning, 2021.
\newblock URL \url{https://arxiv.org/abs/2004.07219}.

\bibitem[Ghugare et~al.(2024)Ghugare, Geist, Berseth, and Eysenbach]{td_sl}
Raj Ghugare, Matthieu Geist, Glen Berseth, and Benjamin Eysenbach.
\newblock Closing the gap between td learning and supervised learning -- a generalisation point of view, 2024.
\newblock URL \url{https://arxiv.org/abs/2401.11237}.

\bibitem[Gupta et~al.(2019)Gupta, Tolani, Davidson, Levine, Sukthankar, and Malik]{congnitive_mapping}
Saurabh Gupta, Varun Tolani, James Davidson, Sergey Levine, Rahul Sukthankar, and Jitendra Malik.
\newblock Cognitive mapping and planning for visual navigation, 2019.
\newblock URL \url{https://arxiv.org/abs/1702.03920}.

\bibitem[Kojima and Deng(2019)]{learn_not_learn}
Noriyuki Kojima and Jia Deng.
\newblock To learn or not to learn: Analyzing the role of learning for navigation in virtual environments, 2019.
\newblock URL \url{https://arxiv.org/abs/1907.11770}.

\bibitem[Mishkin et~al.(2019)Mishkin, Dosovitskiy, and Koltun]{nav_3d}
Dmytro Mishkin, Alexey Dosovitskiy, and Vladlen Koltun.
\newblock Benchmarking classic and learned navigation in complex 3d environments, 2019.
\newblock URL \url{https://arxiv.org/abs/1901.10915}.

\bibitem[OpenAI(2024)]{gpt4}
OpenAI.
\newblock Gpt-4 technical report, 2024.
\newblock URL \url{https://arxiv.org/abs/2303.08774}.

\bibitem[Park et~al.(2025)Park, Frans, Eysenbach, and Levine]{ogbench}
Seohong Park, Kevin Frans, Benjamin Eysenbach, and Sergey Levine.
\newblock Ogbench: Benchmarking offline goal-conditioned rl, 2025.
\newblock URL \url{https://arxiv.org/abs/2410.20092}.

\bibitem[Savva et~al.(2019)Savva, Kadian, Maksymets, Zhao, Wijmans, Jain, Straub, Liu, Koltun, Malik, Parikh, and Batra]{habitat_challenge}
Manolis Savva, Abhishek Kadian, Oleksandr Maksymets, Yili Zhao, Erik Wijmans, Bhavana Jain, Julian Straub, Jia Liu, Vladlen Koltun, Jitendra Malik, Devi Parikh, and Dhruv Batra.
\newblock Habitat: {A} {P}latform for {E}mbodied {AI} {R}esearch.
\newblock In \emph{Proceedings of the IEEE/CVF International Conference on Computer Vision (ICCV)}, 2019.

\bibitem[Schulman et~al.(2017)Schulman, Wolski, Dhariwal, Radford, and Klimov]{ppo}
John Schulman, Filip Wolski, Prafulla Dhariwal, Alec Radford, and Oleg Klimov.
\newblock Proximal policy optimization algorithms, 2017.
\newblock URL \url{https://arxiv.org/abs/1707.06347}.

\bibitem[Wijmans et~al.(2020)Wijmans, Kadian, Morcos, Lee, Essa, Parikh, Savva, and Batra]{ddppo}
Erik Wijmans, Abhishek Kadian, Ari Morcos, Stefan Lee, Irfan Essa, Devi Parikh, Manolis Savva, and Dhruv Batra.
\newblock Dd-ppo: Learning near-perfect pointgoal navigators from 2.5 billion frames, 2020.
\newblock URL \url{https://arxiv.org/abs/1911.00357}.

\bibitem[Xia et~al.(2018)Xia, Zamir, He, Sax, Malik, and Savarese]{gibson}
Fei Xia, Amir Zamir, Zhi-Yang He, Alexander Sax, Jitendra Malik, and Silvio Savarese.
\newblock Gibson env: Real-world perception for embodied agents, 2018.
\newblock URL \url{https://arxiv.org/abs/1808.10654}.

\bibitem[Yang et~al.(2024)Yang, Ding, Brown, Qi, and Xie]{virl}
Jihan Yang, Runyu Ding, Ellis Brown, Xiaojuan Qi, and Saining Xie.
\newblock V-irl: Grounding virtual intelligence in real life, 2024.
\newblock URL \url{https://arxiv.org/abs/2402.03310}.

\end{thebibliography}

\end{document}